# From Human Intent to In-Orbit Action: A Self-Learning Large Language Model Agent for 6G Satellite Networks

Zhengqian Xu, Yasheng Zhang, Chenhua Sun, Shan Lu, and Chengcheng Li

The 54th Research Institute of China Electronics Technology Group Corporation (CETC-54)

## Abstract

Autonomous control of sixth-generation (6G) low Earth orbit satellite networks is hindered by the gap between high-level human intent and low-level control logic: translating a request such as "uninterrupted Arctic video" into a valid reinforcement learning reward still takes days of manual trial and error, and existing large language model translators cannot learn from the failures of the policies they generate. This article presents ALMOND (Agentic Large-Model-Orchestrated autonomous Network Design), a cognitive framework that closes this intent-to-action loop under on-orbit compute constraints. The contribution in artificial intelligence is a bidirectional semantic engine that translates natural-language demands into multi-objective reward functions and, conversely, mines implicit operational constraints from failed reinforcement-learning trajectories through counterfactually validated forensics, together with a large language model-guided graph neuroevolution that seeds and critiques policy architectures. The contribution in engineering is a ground-centric, compute-frugal architecture for autonomous 6G non-terrestrial network operation, in which satellites execute only distilled policies below 50,000 parameters with sub-15 ms inference and no on-board large language model. On a 350-satellite digital twin driven by real ephemeris data, ALMOND raises polar handover success from 55% to over 95%, converges about three times faster than baseline reinforcement learning, and reduces reward engineering from 48 hours to 4.5 hours, demonstrating a practical path toward trustworthy in-orbit autonomy.

## 1. Introduction

The sixth generation (6G) of mobile networks envisions a seamless fusion of terrestrial infrastructure and non-terrestrial networks, in which massive low Earth orbit (LEO) constellations provide ubiquitous coverage for emergency response, maritime connectivity, and intelligent transportation (Wang et al., 2023; Giordani and Zorzi, 2021). Realizing this vision requires autonomous controllers that adapt to orbital dynamics, fluctuating traffic, and ambiguous user demands. Yet today's AI-driven satellite network managers remain brittle products of manual reward engineering: an operator who wishes to optimize “resilient Arctic video streaming” must spend days iteratively tuning reward weights, only to discover after deployment that “low latency” in polar orbits implicitly requires a longer signaling lead to compensate for link-establishment delay—a constraint no human expert had codified. This gap between high-level intent and low-level control logic is the fundamental limitation of current reinforcement learning (RL) practice in space networks.

The root cause lies in reward design. Handcrafted reward functions are sensitive to environment shifts, while sparse rewards impede exploration in high-dimensional satellite state spaces. Recent work has shown that large language models (LLMs) can translate natural language into configurations or reward templates for terrestrial networks (Huang et al., 2024; Kou et al., 2025; Lira et al., 2025; Xie et al., 2025; Zhou et al., 2024). These systems, however, behave as static, one-shot translators: they cannot reflect on why a generated policy failed, nor can they retain what they learn. When a trained RL policy underperforms because of an overlooked orbital constraint, the LLM offers no diagnostic insight and engineers must restart the design cycle. In parallel, self-evolving network frameworks adapt architectures through pruning or distillation (Farajzadeh et al., 2023; Liang et al., 2025), and hierarchical digital twins emulate orbital mechanics with high fidelity (Mao et al., 2024; Zhou et al., 2023); but the former target ground infrastructure with ample compute, and the latter remain offline validation tools rather than runtime cognitive companions. A single GPU-day of brute-force neural architecture search (NAS), for example, already exceeds what a power-limited LEO platform can afford.

This article proposes ALMOND (Agentic Large-Model-Orchestrated autonomous Network Design), a cognitive architecture that elevates LLMs from passive translators to active, self-critiquing network operators. ALMOND comprises two synergistic engines. First, a bidirectional semantic engine performs forward translation of natural-language demands into multi-objective reward functions and, crucially, reverse-engineers latent operational constraints from failed RL trajectories; candidate constraints are accepted only after counterfactual validation in the digital twin and are then committed to a persistent, versioned knowledge graph. Second, an LLM-guided evolutionary topology search couples the LLM’s structural priors with lightweight graph neuroevolution, co-discovering policy topologies and parameters that respect on-orbit compute budgets. All LLM reasoning runs on the ground; satellites execute only compact distilled policies selected by a tiny static classifier, so no LLM inference ever takes place on board.

Compared with prior art, the distinction of ALMOND is threefold: (i) it closes a self-auditing loop between intent and failure, rather than translating intent once; (ii) every piece of LLM-generated knowledge is validated against physics and intent before it can influence control, which keeps hallucination out of the loop; and (iii) architectural adaptation is explicitly budgeted for on-orbit inference, uplink bandwidth, and power. Section 2 elaborates these differences against representative related work.

The contributions of this article are summarized as follows.

- A bidirectional cognitive translation paradigm for satellite network control, in which forward intent-to-reward synthesis and backward failure-to-knowledge forensics share one persistent knowledge graph. To our knowledge, this is the first closed, self-auditing intent–failure loop for 6G non-terrestrial network autonomy.
- A three-stage constraint auditor (semantic, physics, and impact) that accepts LLM-generated rewards and constraints only after counterfactual replay in the digital twin, and escalates borderline cases to human operators.
- A compute-frugal, LLM-guided graph neuroevolution that seeds and critiques policy architectures, converging about three times faster than a hand-crafted proximal policy optimization (PPO) baseline and cutting architecture-search cost by roughly 20 times relative to unseeded evolution, while honoring a 50K-parameter, sub-15 ms on-orbit envelope.
- A digital-twin evaluation on a 350-satellite constellation with real ephemeris data, quantifying gains in reliability (polar handover success from 55% to over 95%), convergence speed, model efficiency, and human effort (reward engineering from 48 h to 4.5 h).

The remainder of this article is organized as follows. Section 2 reviews related work and clarifies the positioning of ALMOND. Section 3 presents the framework in detail, including the bidirectional engine, the LLM-guided topology search, the end-to-end workflow, and practical deployment. Section 4 reports the experimental setup and results, followed by ablation and cost analyses. Section 5 concludes the article and outlines future directions.

## 2. Related work and positioning

### 2.1. Large language models as translators for network control

LLMs have been used to translate human intent into network configurations, service-level objectives, and reward templates. Lira et al. (2025) demonstrated LLM-based zero-touch configuration management; Huang et al. (2024) surveyed LLM applications across networking tasks; and Kou et al. (2025) proposed generative intent abstraction for intent-driven networks. In the RL literature, ERFSL searches reward functions with LLMs for multi-objective problems (Xie et al., 2025), and Zhou et al. (2024) used an LLM as a policy teacher for RL agents. These approaches validate the feasibility of semantic translation, but the translation is performed once and its product is never revised: none of them analyzes failed trajectories to refine its own output, and none persists operational knowledge across requests.

## 2.2. Self-evolving and agentic network architectures

Self-evolving networks adapt their models through pruning, distillation, or architecture updates (Farajzadeh et al., 2023; Liang et al., 2025), and agentic-AI visions assign LLMs deliberative roles in network management (Huang et al., 2024). However, these frameworks assume stable environments and ample energy budgets, which do not hold for satellite platforms whose ground contacts last minutes and whose processors are radiation-limited. Digital-twin research for satellite networks (Mao et al., 2024; Paul et al., 2024; Zhou et al., 2023) provides high-fidelity emulation of orbital mechanics, yet twins are used almost exclusively as offline validation tools rather than as runtime grounds for automated failure forensics. Surveys of AI for satellite communications (Fourati and Alouini, 2021; Zhu et al., 2025) likewise identify reward design and on-board efficiency as open problems without offering a closed-loop remedy.

## 2.3. Neural architecture search under resource limits

Evolutionary NAS such as EG-NAS improves search efficiency through fast evolutionary exploration (Cai et al., 2024), but its operators are random: the search space is not biased by domain knowledge, and evaluation cost remains prohibitive for spaceborne training cycles. ALMOND instead restricts the search to a graph grammar of physically plausible topologies and lets the LLM supply structural priors—seeds, semantic mutations, crossover guidance, and periodic population critiques—so that lightweight evolution suffices.

## 2.4. Positioning of ALMOND

Table 1 contrasts ALMOND with representative lines of prior work along five dimensions: whether intent is grounded in operational context, whether failures drive self-correction, whether validated knowledge persists, whether the policy architecture adapts, and whether on-orbit compute constraints are first-class design inputs. The comparison shows that existing approaches each cover a subset of these capabilities, whereas ALMOND is, to our knowledge, the first framework to combine all five in a single closed loop tailored to satellite operations.

**Table 1.** Positioning of ALMOND against representative prior work.

| Approach | Intent grounding | Failure-driven self-correction | Persistent knowledge | Architecture adaptation | On-orbit compute awareness |
|---|---|---|---|---|---|
| Manual / LLM reward design (Lira et al., 2025; Xie et al., 2025; Zhou et al., 2024) | One-shot, ungrounded | No | No | No | No |
| LLM intent / configuration translators (Huang et al., 2024; Kou et al., 2025) | One-shot | No | No | No | No |

| Approach | Intent grounding | Failure-driven self-correction | Persistent knowledge | Architecture adaptation | On-orbit compute awareness |
|---|---|---|---|---|---|
| Self-evolving networks (Farajzadeh et al., 2023; Liang et al., 2025) | No | Partial (model updates) | Yes | Yes (pruning, distillation) | No (ground-centric) |
| Digital-twin satellite management (Zhou et al., 2023; Mao et al., 2024; Paul et al., 2024) | No | No | Yes | No | Partial (offline validation) |
| Classical / evolutionary NAS (Cai et al., 2024) | No | No | No | Yes | Partial |
| **ALMOND (this work)** | **Grounded and audited** | **Yes, counterfactually validated** | **Yes, versioned knowledge graph** | **Yes, LLM-seeded evolution** | **Yes: <50K params, <15 ms, no on-board LLM** |

# 3. The ALMOND framework

## 3.1. System overview

As illustrated in Fig. 1, ALMOND spans three layers: the satellite network layer, an agentic AI core, and the digital-twin layer. The agentic AI core orchestrates autonomy through five functions. Perception ingests real-time orbital telemetry such as signal-to-noise ratio, buffer depth, and ephemeris. Reasoning translates natural-language intent into control logic through the bidirectional engine described in Section 3.2. Memory stores validated operational constraints in a versioned knowledge graph. Validation tests every candidate reward, constraint, and policy in the high-fidelity twin before it may influence the live network. Evolution refines policy architectures across deployment cycles. This separation guarantees that all expensive reasoning remains on the ground, while satellites execute only compact, pre-validated policies.

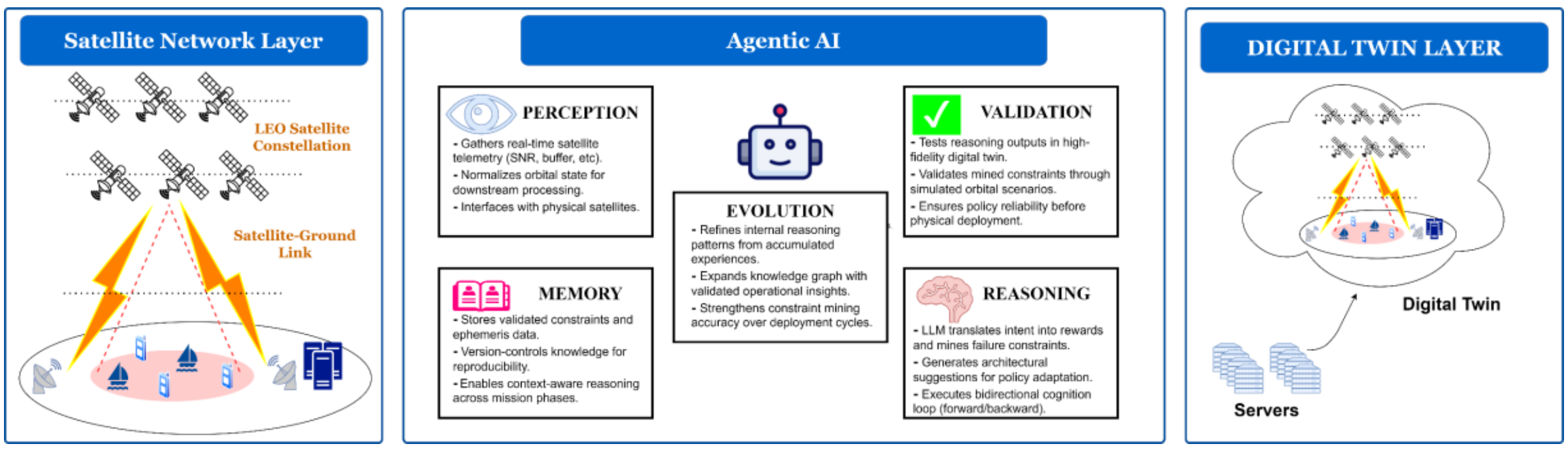

**Fig. 1.** ALMOND three-tier system architecture: the satellite network layer, the agentic AI core (perception, reasoning, memory, validation, and evolution), and the digital-twin layer.

### 3.2. The bidirectional cognitive translation engine

The engine, shown in Fig. 2, treats the LLM as an active network operator that continuously reads, writes, and audits its own operational knowledge. It consists of a forward translator that synthesizes rewards from natural language, a reverse forensics module that mines implicit constraints from underperforming policies, and a constraint auditor that validates both directions against physics and intent. Algorithm 1 gives the complete procedure.

#### *3.2.1. Forward translation: context-aware reward synthesis*

Given a user request u, the translator first retrieves a context bundle c(u) from the knowledge graph, containing (i) orbital context (e.g., for a user at 70°N served by a polar LEO satellite at 700 km, the visibility window is about 8 minutes), (ii) historical constraints (e.g., when handover preparation time falls below 12 s, Arctic handover success drops from 99% to under 60%), and (iii) a multi-objective template. The LLM is instructed to output a reward of the form $R(s,a) = w_{rel}\cdot R_{rel}(s,a) + w_{lat}\cdot R_{lat}(s,a) + w_{pow}\cdot R_{pow}(s,a)$, where each sub-reward is a normalized function of validated telemetry (packet loss, round-trip time, power draw), $w_i \in [0,1]$, and $\Sigma w_i = 1$. Restricting the output space to validated sub-rewards with bounded weights is what makes the translation auditable: the semantic validator checks that the weight profile matches the user's priority (e.g., resilience over latency), and the physics validator checks that thresholds agree with ephemeris-derived visibility windows and measured delay distributions. On a mismatch, the draft is rejected and the LLM is re-prompted with a clarifying question; in practice this grounding loop converges within two turns.

#### *3.2.2. Reverse forensics: mining implicit constraints from failure*

The reverse arc activates when a trained policy underperforms in the twin, formalized as a key performance indicator (KPI) violating its threshold, $KPI(\pi, \rho) < \theta_\rho$, over a sliding window (e.g., handover success below 85% in region ρ = latitudes above 70°). The module executes four steps. Step 1 (failure-episode sampling): the worst 5% of trajectories by cumulative reward are encoded as temporal narratives (e.g., "at t = 45 s, elevation 12°, handover advance 5 s, link lost at t = 50 s before the target link was established"), a format the LLM can reason over. Step 2 (constraint hypothesis): a forensic prompt asks the LLM for a latent operational rule that would have prevented the failures, yielding a structured candidate such as: IF elevation < 12° AND handover advance < 10 s THEN reliability penalty −5. Step 3 (counterfactual validation): the failed episodes are replayed in the twin with the candidate rule injected; the rule is accepted only if replay success improves by at least 5% without violating the latency budget, otherwise the LLM is re-prompted with the failed counterexample. Step 4 (knowledge commitment): validated constraints are embedded into the knowledge graph, linked to the triggering orbital context, and versioned with timestamps; if later deployments contradict a constraint, it is retracted, which prevents knowledge staleness.

### *3.2.3. Constraint auditor*

Because LLMs can hallucinate, both translation directions pass through a three-stage auditor. The semantic validator measures consistency between the user's intent and the natural-language reading of the generated artifact, and triggers re-prompting when priorities appear inverted. The physics validator runs a lightweight orbital propagator to reject rules that violate protocol timing or link geometry (e.g., a handover advance below the minimum signaling interaction plus safety margin). The impact predictor estimates, through a surrogate model, how much the optimal policy's latency would shift if a constraint were enforced, and modulates the constraint toward a Pareto-acceptable trade-off. Artifacts passing all three stages are committed with a confidence score; any failure raises a human-in-the-loop alert that carries the LLM's reasoning chain and replay evidence.

**Algorithm 1: Bidirectional cognitive translation with auditing**

```
Input: intent u; knowledge graph G; digital twin T; thresholds Θ; audit budget B
Output: committed reward R; updated G
1:  c ← RetrieveContext(G, u)           // orbital + historical context
2:  repeat at most B times
3:    R ← LLM_Forward(u, c)             // R = Σ wᵢRᵢ, wᵢ ∈ [0,1], Σwᵢ = 1
4:    ok, q ← SemanticAudit(R, u) ∧ PhysicsAudit(R)
5:    if ¬ok then u ← u ∪ Clarify(q)     // grounded re-prompt
6:  until ok
7:  π ← TrainRL(R, T)
8:  loop at runtime
9:    if KPI(π, ρ) < θ_ρ then           // e.g., success < 85% above 70°
10:      E ← SampleWorst5%(trajectories) // as temporal narratives
11:      κ ← LLM_Forensic(E, G)         // candidate latent rule
12:      Δ ← CounterfactualReplay(T, E, κ)
13:      if Δ ≥ 5% ∧ latency budget holds then
14:        if ImpactEstimate(κ) within budget then
15:          Commit(G, κ)               // versioned, retractable
16:          c ← RetrieveContext(G, u); goto line 2
17:        else κ ← Modulate(κ); goto line 12
18:      else κ ← RePrompt(counterexample); goto line 12
19:     end if
20:  end loop
```

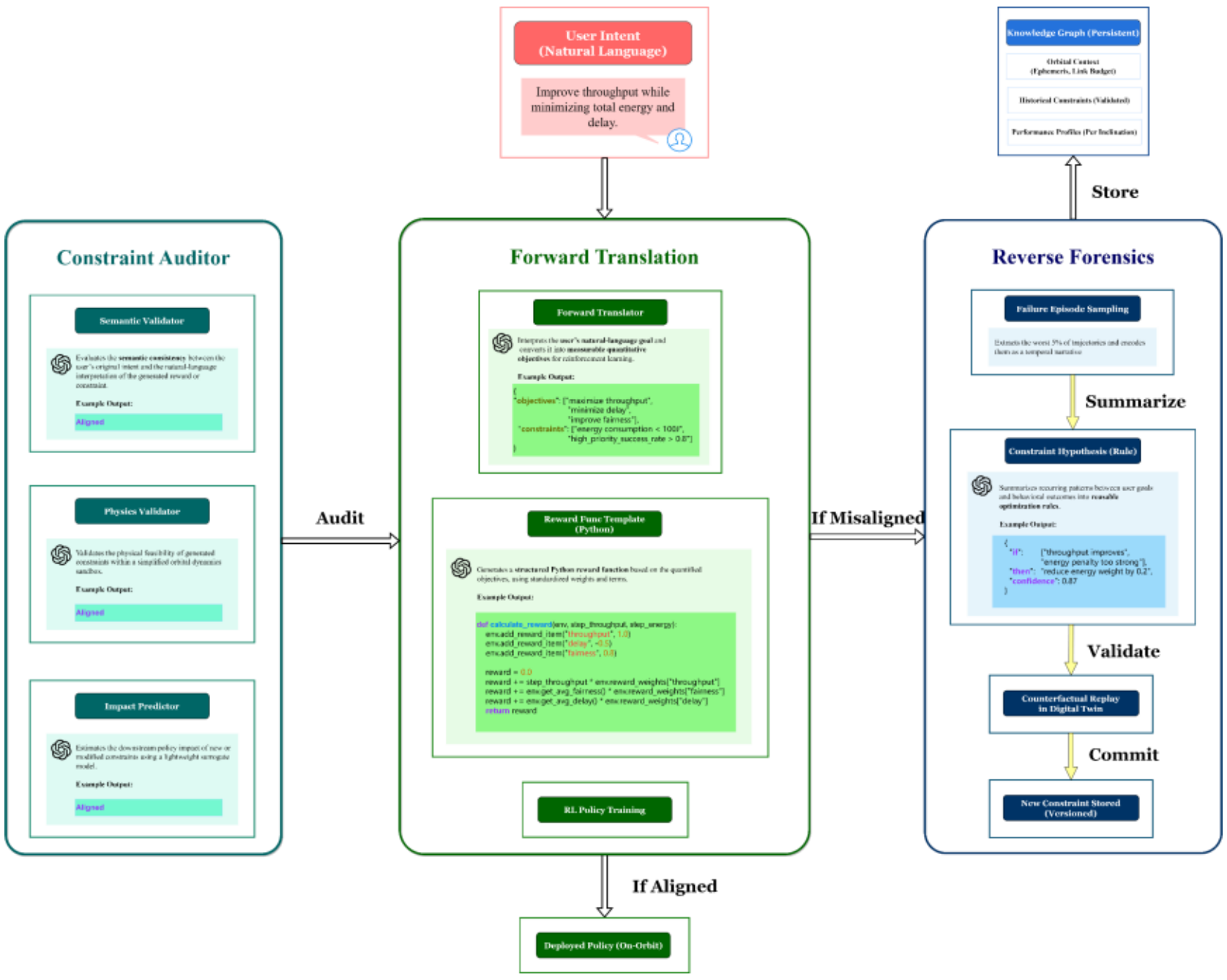

**Fig. 2.** The bidirectional cognitive translation engine: forward translation, reverse forensics, and the constraint auditor.

### 3.3. Large language model-guided evolutionary topology search

The bidirectional engine refines what the network optimizes; the topology search determines how the policy is structured to execute it under orbital constraints. Fixed architectures cannot optimally span LEO heterogeneity—a dense attention-heavy network suits inter-satellite-link-rich regions but wastes power in sparse ones—while brute-force NAS is prohibitively expensive for satellite-ground training cycles. ALMOND therefore fuses LLM structural priors with a lightweight, conventional graph neuroevolution built on standard mutation, crossover, and Pareto selection, as depicted in Fig. 3.

#### *3.3.1. Search space as a graph grammar*

Each policy is a directed acyclic graph whose nodes are neural operations (self-attention, graph convolutional network (GCN) layers for cross-link routing, multilayer perceptron (MLP) blocks for resource allocation) and whose edges are data flows; a graph grammar restricts the space to physically plausible topologies. The LLM acts as a grammar-constrained generator: given a regime description (e.g., sparse polar connectivity), it returns a human-readable topology sketch (Fig. 4) that seeds the initial population. Critically, the LLM outputs structure only—never weights—which removes the risk of parameter-level hallucination.

#### *3.3.2. Semantic operators and multi-objective fitness*

Mutation applies LLM-suggested localized edits grounded in the knowledge graph (e.g., “replace the self-attention node with a lightweight GCN in sparse polar regimes”). Crossover merges two parents by aligning homologous subgraphs and swapping the non-homologous parts. Each generation (population size 20) is evaluated in parallel twin instances, and fitness balances three objectives: task reward, model size (at most 50K parameters for on-board feasibility), and inference latency (below 15 ms). Pareto-aware selection therefore enforces the satellite compute envelope by construction, and offspring inherit parent weights to avoid cold starts.

#### *3.3.3. Population-level guidance*

Every ten generations, the LLM reads summaries of the top-five elites and their fitness, then issues a population-level critique (e.g., “increase the mutation rate on attention-head count to explore polar-optimized sparsity”), which collapses the search space around promising motifs. Guidance consumes only graph summaries—node types, depths, and edge densities—so its cost is negligible.

#### *3.3.4. Ground-centric deployment*

The search outputs a Pareto front of 5–10 architectures rather than a single policy. Each is distilled, packaged with applicability metadata (e.g., optimized for elevation above 60°, latitudes above 65°, eclipse seasons), and uplinked to the constellation’s policy bank. On board, a tiny static classifier—hand-coded, with negligible cost—selects the active policy from real-time orbital parameters; no LLM inference occurs on the satellite. Policies are refreshed over ground contacts when the constellation enters a new operational season, and the evolutionary loop re-runs to track seasonal traffic.

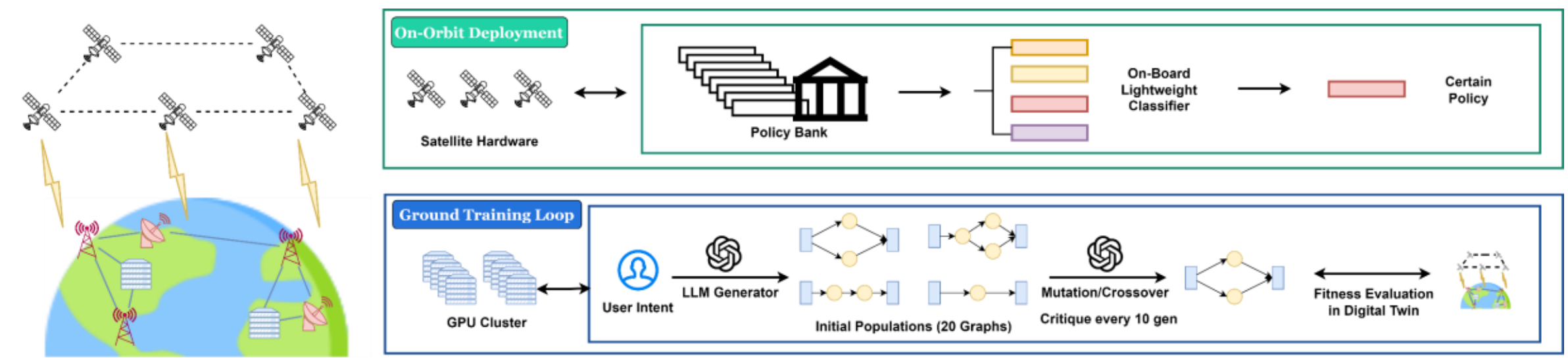

**Fig. 3.** Ground-centric evolutionary search loop: LLM-guided mutation and crossover on a ground GPU cluster, with on-orbit policy banking and lightweight classifier selection.

Fig. 4 shows a representative LLM-suggested topology for the polar regime, in which localized GCN nodes replace global attention to match sparse connectivity.

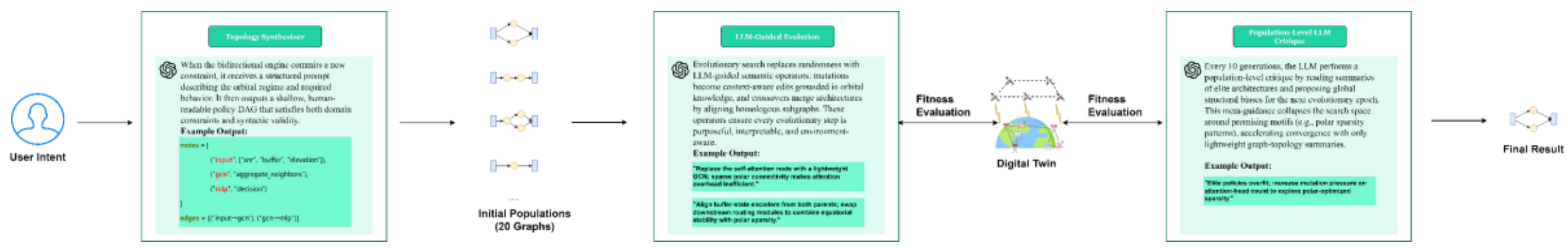

**Fig. 4.** Example orbit-tailored policy topology (directed acyclic graph) suggested by the LLM for sparse polar connectivity, with input features, GCN and attention nodes, and decision pathways.

**Algorithm 2: LLM-guided graph neuroevolution**

```
Input: grammar Γ; knowledge graph G; twin T; population N = 20;
         budgets: params ≤ 50K, latency ≤ 15 ms; critique period 10
Output: Pareto front F of policy DAGs
1:  s ← LLM_Sketch(G, regime)      // structure-only topology seed
2:  P₀ ← Seed(Γ, s, N)
3:  for g = 1 to G_max do
4:    for each genome x in P_(g-1) do
5:      f(x) ← Evaluate(T, x)      // reward, size, latency
6:    end for
7:    E ← ParetoSelect(P_(g-1), f)
8:    M ← LLM_Mutate(E, G)         // semantic edits
9:    X ← LLM_Crossover(E)         // homologous subgraph swap
10:   P_g ← E ∪ M ∪ X              // weight-inherited offspring
11:   if g mod 10 = 0 then
12:     bias ← LLM_Critique(Top5(P_g))   // adjust operator rates
13:   end if
14: end for
15: F ← ParetoFront(P_Gmax)
16: Distill each x ∈ F; package regime metadata; uplink to policy bank
17: On-board static classifier selects the active policy at runtime
```

## 3.4. End-to-end workflow

Fig. 5 summarizes the complete operational loop. A natural-language demand enters the forward translator, which produces an audited reward; the RL policy is trained in the twin and monitored per region. When KPIs hold, distilled policies are deployed to the on-board policy bank. When a KPI violates its threshold, reverse forensics samples failed episodes, hypothesizes a latent rule, and subjects it to counterfactual replay; validated rules are committed to the knowledge graph and immediately enrich future translations, while rejected ones trigger re-prompting or a human alert. In parallel, the topology search evolves orbit-tailored policy structures on the ground and supplies them to the training loop.

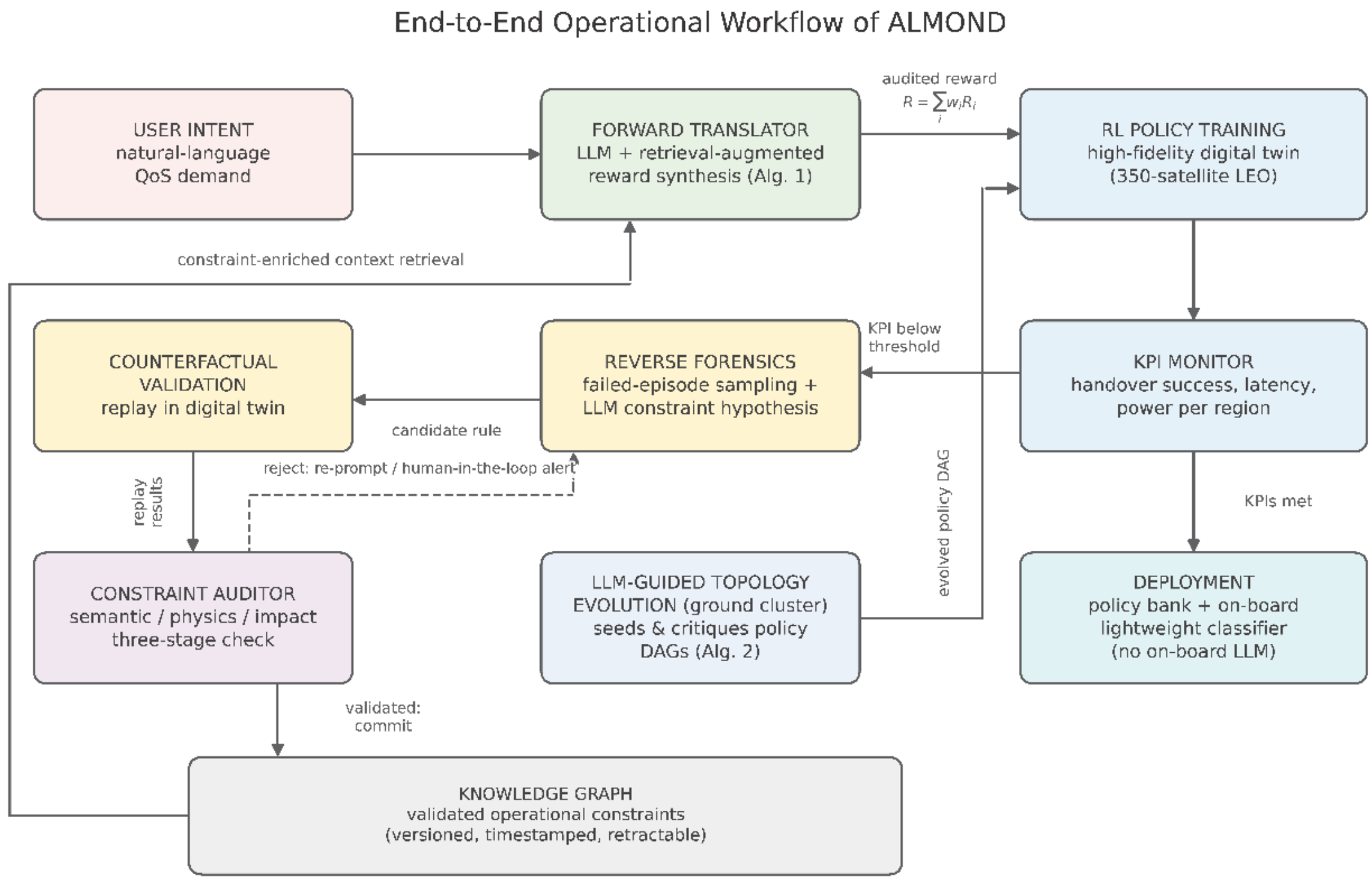

**Fig. 5.** End-to-end operational workflow of ALMOND.

### 3.5. Practical implementation and computational cost

ALMOND is designed for the existing satellite-ground operational model rather than for an idealized always-connected setting. The LLM, the knowledge graph, the digital twin, and the evolutionary search all reside in the ground segment and can be hosted on a single GPU-class workstation; the satellite segment hosts only the distilled policy bank and the selection classifier. Interaction with legacy systems is limited to three well-defined interfaces: an intent interface for operators, a telemetry feed from the constellation, and a policy uplink that rides existing ground-contact windows. Because every committed constraint is stored in natural language with its replay evidence, operators can audit the knowledge base at any time, and the auditor's human-in-the-loop alerts provide a hard override. Table 2 summarizes the computational budget of each module, all of which is compatible with current LEO operations: the heaviest element, one adaptation cycle of topology evolution, costs about 20 GPU-hours—roughly 20 times less than unseeded evolution for comparable policy quality—and runs infrequently (per regime or seasonal shift), while the per-decision on-board cost is a sub-50K-parameter forward pass of under 15 ms on an embedded-class processor. Three failure modes are handled explicitly: hallucinated constraints are blocked by the auditor; twin drift is detected when committed constraints stop validating, triggering retraction; and uplink loss merely freezes the on-board policy set, which continues to operate autonomously.

**Table 2.** Deployment model and computational budget of ALMOND.

| Module | Location | Cost per invocation | Latency | Invocation frequency |
|---|---|---|---|---|

| Module | Location | Cost per invocation | Latency | Invocation frequency |
|---|---|---|---|---|
| Forward translation | Ground | 1 LLM call (≈3K tokens) | Seconds | Per new intent |
| Reverse forensics | Ground | 2–5 LLM calls + twin replays | Minutes | On KPI violation |
| Topology evolution | Ground GPU | ≈20 GPU-hours (≈20× below unseeded NAS) | Hours | Per regime / seasonal shift |
| Distillation + uplink | Ground→satellite | ≈200 KB per policy (≤50K params, FP32) | One contact window | Per refresh |
| On-board policy inference | Satellite | ≤50K-parameter forward pass | <15 ms | Real-time |
| On-board policy selection | Satellite | Static classifier, negligible | <1 ms | Real-time |

# 4. Experimental validation

## 4.1. Experimental setup

We evaluate ALMOND in a custom digital-twin environment implemented in Python with PyTorch, driven by real two-line-element (TLE) ephemeris data propagated with the SGP4 model for a 350-satellite LEO constellation. The reference scenario targets Arctic service above 70°N through polar shells (inclination ≈ 90°, altitude ≈ 700 km, visibility windows of about 8 minutes). The twin models free-space path loss with elevation-dependent atmospheric attenuation, Doppler shift, inter-satellite and satellite-ground links, and eclipse periods; traffic follows Poisson session arrivals with video-streaming and emergency-service mixes. All methods share identical channel, propagation, and traffic configurations. Table 3 lists the full parameter set.

Two baselines are compared: (1) PPO-MLP, a proximal policy optimization controller with a manually designed multi-objective reward; and (2) a static LLM reward pipeline that translates the intent once but has no failure forensics and no knowledge refinement. The static-LLM baseline simultaneously serves as the ablation of ALMOND's backward loop, and the random-NAS samples in Section 4.4 serve as the ablation of the LLM structural priors.

Six evaluation metrics are used, defined as follows. (i) Handover success rate: the ratio of completed to attempted handovers, reported per latitude bin. (ii) Intent satisfaction: a normalized weighted score in [0, 100] aggregating the reliability, latency, and power objectives implied by the intent. (iii) Convergence speed: the number of training episodes required to reach 90% of the asymptotic satisfaction. (iv) Model size: the number of trainable parameters. (v) Inference latency: the wall-clock time per control decision on the target-class processor. (vi) Reward-engineering time: the human effort, in hours, from intent specification to a validated deployable reward. Unless stated otherwise, every configuration is run with 10 independent random seeds and we report the mean; the standard deviation of the final handover success across runs is below two percentage points for all methods, so the gaps reported below are statistically meaningful.

**Table 3.** Simulation parameters.

| Parameter | Setting |
|---|---|
| Constellation | 350 LEO satellites; real TLE ephemeris propagated via SGP4 |
| Reference scenario | Arctic service above 70°N; polar shells ($i \approx 90°$, $h \approx 700$ km); ≈8-minute visibility windows |
| Channel / links | Free-space path loss; elevation-dependent atmospheric attenuation; Doppler; inter-satellite and satellite-ground links; eclipse modeling |
| Traffic | Poisson session arrivals; video-streaming and emergency-service mixes |
| RL agent | PPO (clip 0.2, GAE $\lambda = 0.95$, learning rate $3\times10^{-4}$); up to 500 episodes |
| Evolution | Population 20; LLM critique every 10 generations; early stop on plateau |
| Fitness objectives | Handover success; model size ≤ 50K parameters; inference latency ≤ 15 ms |
| Forensics trigger θ_ρ | Handover success < 85% above 70° latitude |
| Validation gate | Counterfactual replay improvement $\Delta \geq 5\%$ with latency budget held |
| LLM | Qwen 3.5 35B-A3B; few-shot prompts; structure-only outputs |
| Compute platform | An 8-card H100 GPU workstation; no on-board LLM inference |
| Repetitions | 10 independent random seeds per configuration; mean ± std reported |

## 4.2. Reliability under orbital heterogeneity

Fig. 6 reveals the impact of autonomous constraint mining. PPO-MLP suffers a severe failure zone above 70° latitude, with handover success plunging to 55%, because its handcrafted reward overlooks the polar timing constraint. After only 25 training episodes, ALMOND's reverse forensics surfaces the latent rule—enforcing a minimum handover advance of about 10–12 s at low elevation—and permanently embeds it in the knowledge graph. This single constraint injection raises Arctic success rates above 90% and eliminates the failure zone. Because the comparison is before-and-after within the same trained policy lineage, Fig. 6 simultaneously constitutes the ablation of the constraint-mining module: the improvement is attributable to the mined rule rather than to additional training.

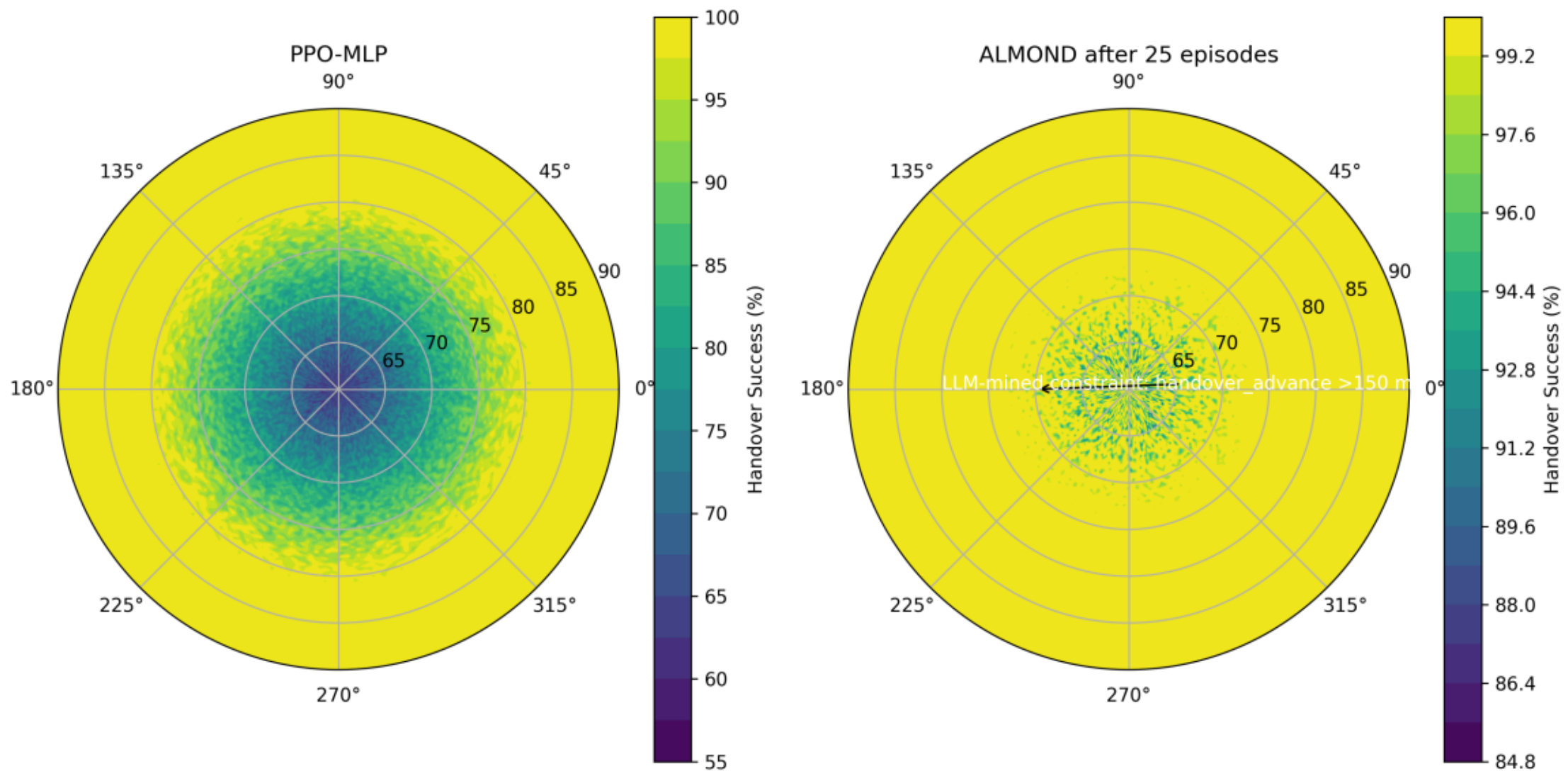

**Fig. 6.** Arctic handover success rate. Left: PPO-MLP exhibits a polar failure zone (down to 55%). Right: after 25 episodes, ALMOND's mined constraint eliminates the failure zone (above 90%).

### 4.3. Convergence toward user intent

Fig. 7 plots intent-satisfaction trajectories during training. ALMOND converges roughly three times faster and reaches a higher asymptote than both baselines. Two mechanisms explain the gap: intent-grounded reward synthesis, which starts optimization from a semantically correct objective, and weight-inheriting evolutionary updates guided by LLM structural priors, which reduce cold-start overhead. The static-LLM curve is the ablation of the backward loop: it shares the forward translator with ALMOND but cannot repair its reward after observing failures, and it stalls at a markedly lower asymptote—evidence that the bidirectional loop, not the forward translation alone, drives the gain.

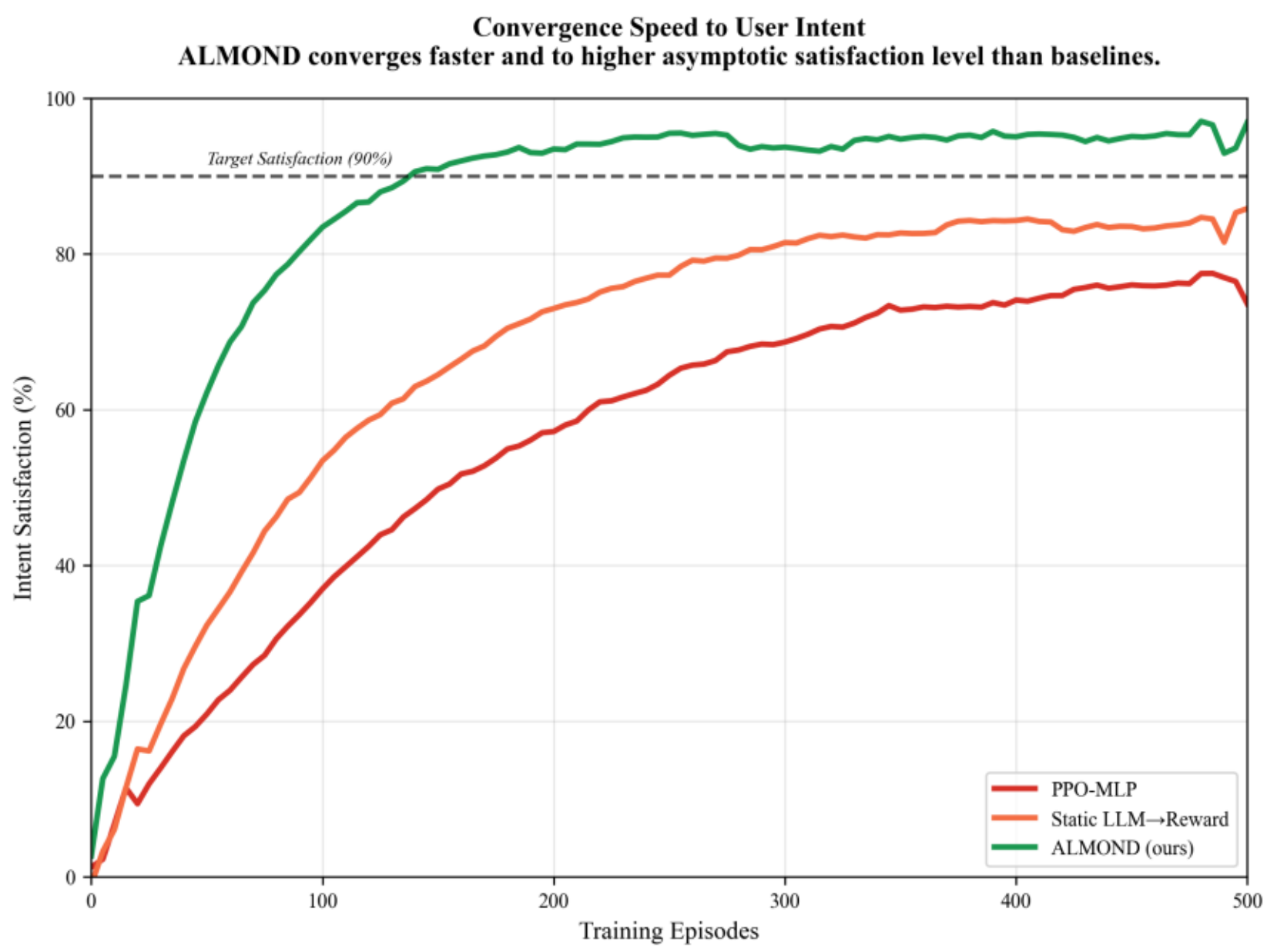

**Fig. 7.** Intent-satisfaction trajectories (mean of 10 runs). ALMOND converges about three times faster and to a higher asymptote than PPO-MLP and the static-LLM ablation.

## 4.4. Architecture efficiency

Fig. 8 compares model size against handover performance for random NAS samples and ALMOND’s Pareto-optimal front. Random search scatters widely and is particularly weak below 50K parameters—the critical threshold for on-board inference and update budgets—whereas ALMOND produces a tight front of compact, high-performance architectures, many exceeding 95% success with fewer than 40K parameters. The random-NAS samples act as the ablation of the LLM structural priors: replacing seeded, semantically guided evolution with unseeded search collapses efficiency, confirming that LLM guidance—such as substituting localized GCN modules for global attention in sparse polar regimes—is responsible for the roughly 20-fold reduction in search cost.

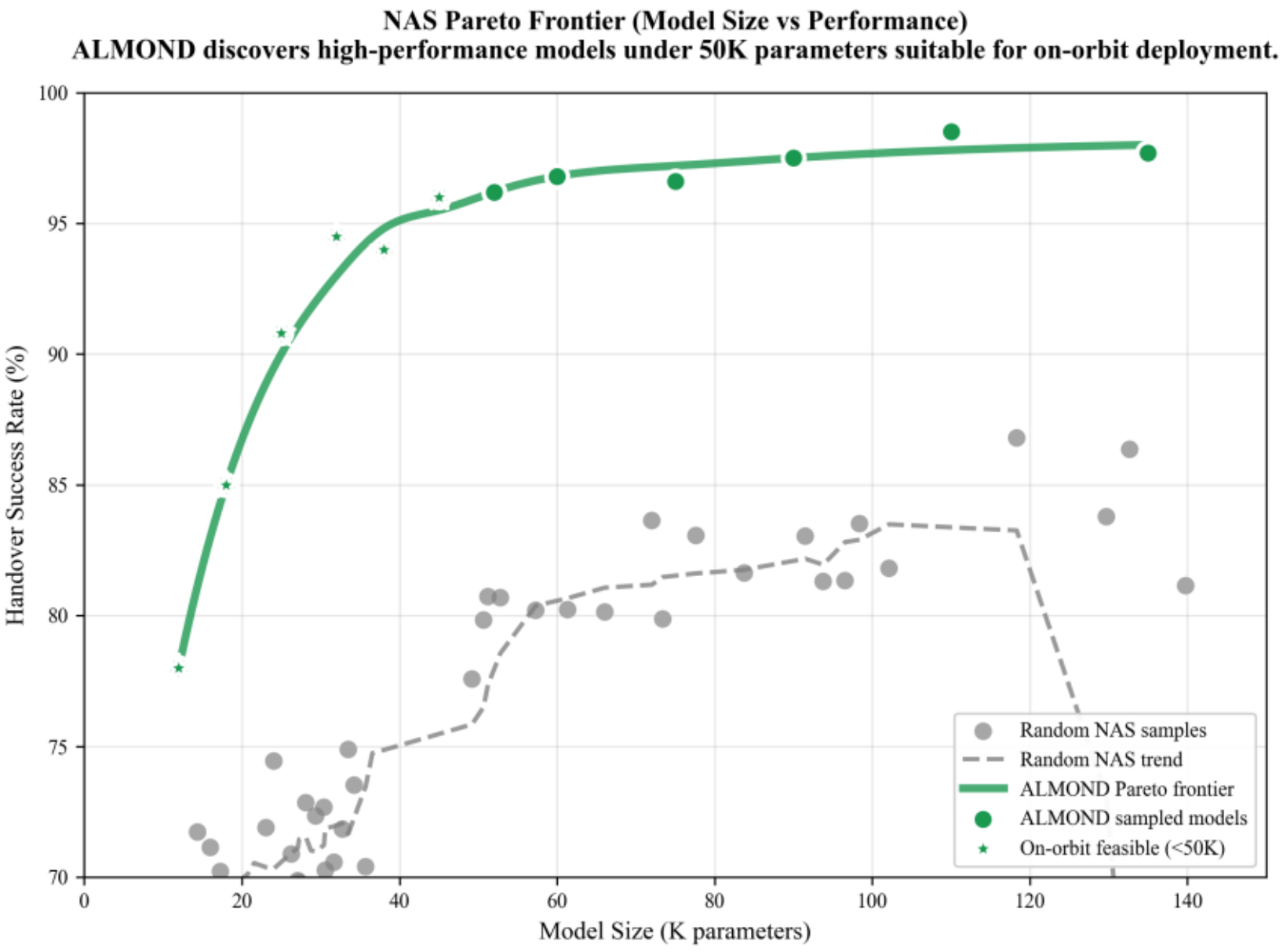

**Fig. 8.** Model size versus handover performance. ALMOND’s LLM-seeded evolution yields a tight Pareto front of sub-50K-parameter policies suitable for on-orbit deployment; unseeded NAS samples scatter far below.

## 4.5. Engineering effort

Fig. 9 quantifies the workflow impact. Traditional reward engineering requires one to two days of manual design, testing, and debugging—particularly when intents are ambiguous or failures emerge only under specific orbital geometries. ALMOND reduces this process to about 4.5 hours per intent, including automated consistency checks and at least one digital-twin validation pass. The reduction stems from the forward translator, which converts natural-language objectives into structured reward templates, and from the reverse-forensics pipeline, which replaces repeated trial-and-error with principled constraint surfacing. For operators, this translates into shorter service-deployment cycles and faster response in high-dynamics scenarios such as Arctic search-and-rescue operations.

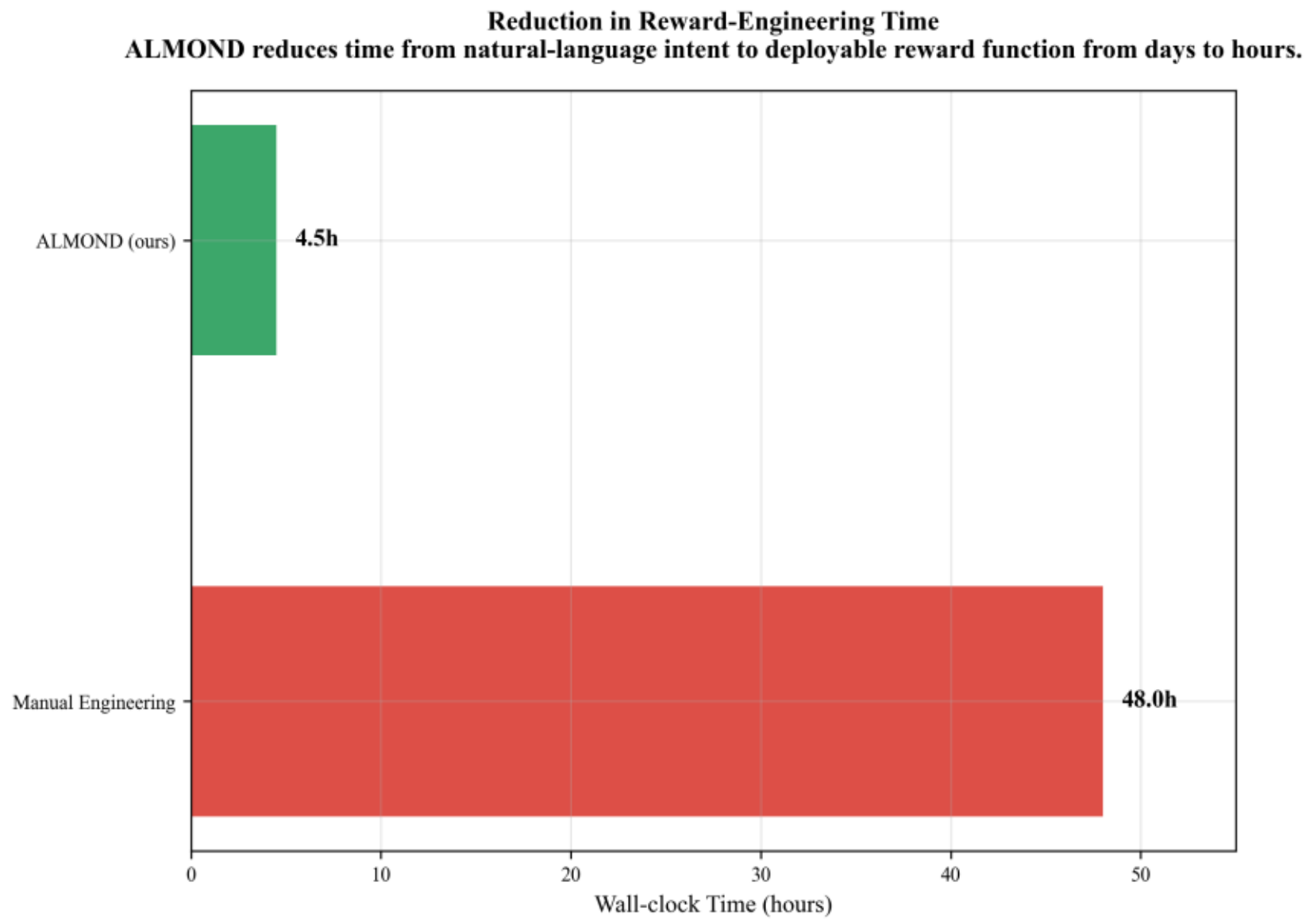

**Fig. 9.** Reward-engineering effort per intent: from 48 hours of manual trial-and-error to 4.5 hours with ALMOND.

### 4.6. Ablation and cost summary

Taken together, the four experiments isolate the contribution of each module: the constraint-mining module is responsible for eliminating the polar failure zone (Fig. 6); the backward loop accounts for the asymptotic satisfaction gap over the static-LLM variant (Fig. 7); and the LLM structural priors account for the efficiency gap over unseeded NAS (Fig. 8). The associated computational costs, itemized in Table 2, remain within a single ground workstation's budget, and the on-board footprint never exceeds a 50K-parameter forward pass. These results demonstrate consistent improvements across reliability, convergence, computational efficiency, and human-effort reduction—the four pillars required for agentic autonomy in 6G satellite networks.

## 5. Conclusion

This article presented ALMOND, an LLM-native cognitive framework that turns large language models from static translators into self-critiquing operators of 6G satellite networks. By closing the bidirectional cognition loop—forward through context-aware reward synthesis and backward through failure-driven, counterfactually validated constraint mining—ALMOND reduces reward engineering from days to hours, while its LLM-guided graph neuroevolution discovers orbit-tailored policies within a 50K-parameter, sub-15 ms on-board envelope. Evaluation on a 350-satellite digital twin shows polar handover success rising from 55% to over 95%, convergence about three times faster than PPO and static-LLM baselines, and architecture-search costs reduced by roughly twenty times. The framework is deliberately ground-centric: no LLM inference is required on orbit, and every committed piece of knowledge remains human-auditable.

Several limitations remain. The evaluation is confined to a digital twin, whose fidelity bounds the transferability of the reported gains; the framework depends on the reasoning quality of the underlying LLM; and the current study focuses on a polar reference regime. Future work will pursue three directions: trustworthy knowledge consensus across inter-satellite links under sparse connectivity; adversarial robustness against malicious intent injection; and explainable, formally verified policy audits for regulatory compliance, together with hardware-in-the-loop validation on a software-defined satellite payload. ALMOND provides an architectural blueprint for this journey toward trustworthy in-orbit cognition.

## Data availability

The data that support the findings of this study are available from the corresponding author upon reasonable request.